\documentclass[letterpaper, 10 pt, conference]{ieeeconf}  % Comment this line out if you need a4paper

\IEEEoverridecommandlockouts                              % This command is only needed if 
\usepackage{multirow}  
\usepackage{multicol}
\usepackage{array,multirow,graphicx}
\usepackage[table]{xcolor}
\usepackage{tikz}
\usepackage{hhline}
\newcommand\copyrighttext{\footnotesize \textcopyright~2023 IEEE. Personal use of this material is permitted. Permission from IEEE must be obtained for all other uses, in any current or future media, including reprinting/republishing this material for advertising or promotional purposes, creating new collective works, for resale or redistribution to servers or lists, or reuse of any copyrighted component of this work in other works.
}%

\newcommand\copyrightnotice{%
	\begin{tikzpicture}[remember picture,overlay]
	\node[anchor=south,xshift=0pt,yshift=14pt] at (current page.south) {\fbox{\parbox{\dimexpr\textwidth-\fboxsep-\fboxrule\relax}{\copyrighttext}}};
	\end{tikzpicture}%
}% Needed to meet printer requirements.

\title{\LARGE \bf
Augmentation-based Domain Generalization for Semantic Segmentation
}

\author{Manuel Schwonberg$^{1,2}$, Fadoua El Bouazati$^{2}$, Nico M. Schmidt$^{1}$ and Hanno Gottschalk$^{2}$% <-this % stops a space
\thanks{$^{1}$ CARIAD SE,
        {\tt\small firstname.lastname@cariad.technology}}%
\thanks{$^{2}$ IZMD \& School of Mathematics and Natural Sciences, University of Wuppertal,
        {\tt\small firstname.lastname@uni-wuppertal.de}}%
}

\begin{document}

\maketitle
\thispagestyle{empty}
\pagestyle{empty}

%%%%%%%%%%%%%%%%%%%%%%%%%%%%%%%%%%%%%%%%%%%%%%%%%%%%%%%%%%%%%%%%%%%%%%%%%%%%%%%%
\begin{abstract}

Unsupervised Domain Adaptation (UDA) and domain generalization (DG) are two research areas that aim to tackle the lack of generalization of Deep Neural Networks (DNNs) towards unseen domains. While UDA methods have access to unlabeled target images, domain generalization does not involve any target data and only learns generalized features from a source domain. Image-style randomization or augmentation is a popular approach to improve network generalization without access to the target domain. Complex methods are often proposed that disregard the potential of simple image augmentations for out-of-domain generalization. For this reason, we systematically study the in- and out-of-domain generalization capabilities of simple, rule-based image augmentations like blur, noise, color jitter and many more. Based on a full factorial design of experiment design we provide a systematic statistical evaluation of augmentations and their interactions. Our analysis provides both, expected and unexpected, outcomes. Expected, because our experiments confirm the common scientific standard that combination of multiple different augmentations outperforms single augmentations. Unexpected, because combined augmentations perform competitive to state-of-the-art domain generalization approaches, while being significantly simpler and without training overhead. On the challenging synthetic-to-real domain shift between Synthia and Cityscapes we reach 39.5\% mIoU compared to 40.9\% mIoU of the best previous work. When additionally employing the recent vision transformer architecture DAFormer we outperform these benchmarks with a performance of 44.2\% mIoU.        
\end{abstract}

%%%%%%%%%%%%%%%%%%%%%%%%%%%%%%%%%%%%%%%%%%%%%%%%%%%%%%%%%%%%%%%%%%%%%%%%%%%%%%%%
\section{Introduction}
\copyrightnotice
Deep Neural Networks (DNNs) have shown remarkable capabilities for a wide range of tasks in the past years. In particular,  classification, object detection and semantic segmentation emerged as dynamically evolving research areas. Camera-based semantic segmentation is a crucial part of the perception system of autonomous vehicles to obtain a fine-grained understanding of the environment \cite{fingscheidt_dnndataautomateddriving}.\\
However, major challenges for a robust application of DNNs in autonomous vehicles remain to be solved. The lack of generalization to new, unseen domains can be a serious threat for the safe perception \cite{Lohdefink_2020_CVPR_Workshops}. In operation in the real world, an autonomous vehicle will encounter several domain shifts. Day-to-night shift, varying weather conditions and different locations are some of the most relevant shifts the perception system has to handle. The utilization of synthetic data, where labels are available without manual annotation effort, for model validation and the simulation of rare and/or dangerous driving scenarios, is also impeded.\\ 
Domain shifts cause severe drops of performance,  when DNNs are employed \cite{pmlr-v205-niemeijer23a}. This hinders large-scale application in autonomous driving. In the past years, an active research area has emerged which aims to mitigate the domain shift and reduce the performance drop. On a high-level, we distinguish between domain adaptation (DA) and domain generalization (DG) methods. Domain Adaptation refers to the setting where labels in the source domain are available and for the target domain no or only a subset of labels is accessible \cite{Hoffman2016}. In particular unsupervised domain adaptation has gained a lot of  attention in research and recent works like ProDA \cite{zhang2021prototypical}, SePiCo \cite{xie2022sepico} and DAFormer \cite{hoyer2022daformer} with vision transformers achieved remarkable results for semantic segmentation. However, any form of domain adaptation requires access to target data (without labels) which is difficult to achieve, because the target domain often is unknown in advance \cite{kim2021wedge}. Therefore, the field of DG assumes that during training source data is available, but no target data. In line with the observation from \cite{huang2021fsdr}, we find a significantly smaller number of approaches covering DG than for UDA. \\
We identify three groups of DG approaches for semantic segmentation: domain randomization without external real data, domain randomization using external real data and  instance/batch whitening and normalization. Augmentations like color jittering, noise, blur etc.\ are commonly used to improve in-domain generalization, but can also be seen as image randomization. However, there exist no works either in  DA nor DG systematically investigating the potential of augmentations for out-of-domain generalization.\\ %Augmentations are applicable to both adaptation and generalization and can be integrated into any training process without changes of the actual training pipeline. 
This gives us a strong motivation to systematically evaluate the potential of augmentations for DG for semantic segmentation. In the present work, we not only provide a quantitative analysis, in as much various augmentations improve domain generalization, but also provide a systematic evaluation that enables researchers to employ augmentations in a more targeted manner. Our contributions are:
\begin{itemize}
    \item we evaluate an extensive set of augmentations on the problem of synthetic-to-real domain generalization; 
    \item we employ both a ResNet-101 based architecture and as the very first work for DG the novel vision transformer DAFormer \cite{hoyer2022daformer} architecture for evaluation. This is the first time to the best of our knowledge that vision transformers are benchmarked for domain generalization for semantic segmentation; 
    \item we evaluate the best performing set of augmentations in a full factorial manner and statistically analyze their interactions;
    \item our comparison with state-of-the-art DG approaches reveals that simple augmentations can provide competitive performance with more complex approaches. 
\end{itemize}

\section{Related Work}
\textbf{Unsupervised Domain Adaptation (UDA)} In this setting unlabeled target data is available to adapt the model for the target domain. Building on this, a large variety of UDA approaches have been proposed. Similar to DG, several UDA approaches operate in the input space and propose to align the domains by style transfer. These methods aim to transfer the visual appearance of one domain to the other domain by using GAN-based techniques \cite{sankaranarayanan2018learning, chen2019learning, chang2019all} and often employ a CycleGAN \cite{Zhu2017} for a cycle-consistent style transfer \cite{lee2018spigan, Chen2019crdoco, yang2021context}. Image Augmentations are utilized for UDA by consistency learning \cite{araslanov2021self} and to learn texture invariant representations \cite{kim2020learning}. Next to the input space UDA, further methods operate in the feature and output space. Also in these spaces, image augmentations on the input-level are an essential prerequisite. For contrastive learning, augmentations can generate two different versions of the same image. Thereby, they generate positive and negative pairs for the contrastive loss \cite{zhou2021domain, shim2021learning}. Similarly, augmentations support self-training with knowledge distillation where a teacher network usually receives a non-augmented or weakly augmented and the student a strongly augmented image combining self-training with consistency learning \cite{zhang2022unsupervised, xie2022sepico, gao2021dsp, zhou2021context}. Augmentations in UDA are mostly used as a helper tool so that their standalone performance remains unclear.    
\\
\textbf{Domain Generalization (DG)}
In domain generalization, no target data is available and also the objective differs since researchers aim to train models which generalize to multiple unseen domains. As mentioned, DG approaches can be categorized into three different groups: instance/batch normalization and whitening methods, domain randomization with and without accessing real data. Belonging to the first category IBN-Net \cite{pan2018two} was one of the first DG approaches for semantic segmentation. The authors propose architecture-level changes for the location of the instance (IN) and batch normalization (BN) layers based on the assumption that IN helps learning style-invariant and BN content-specific features. RobustNet \cite{choi2021robustnet} uses this architecture as a basis to reduce the covariance of style and appearance-dependent features with an instance selective whitening loss. The work from Pan et al. \cite{pan2019switchable} extends the idea of fixed whitening by trainable importance weights that the network can use to switch in a task-dependent way between whitening and normalization. Their results show that the importance weights strongly differs between different tasks. SAN+SAW \cite{peng2022semantic} is the latest approach in the line of whitening and normalization DG methods. The authors argue that other methods do not distinguish between different classes and propose semantic-aware normalization (SAN) and whitening (SAW).\\
The second large group in DG is domain randomization with additional real data that does not belong to the unseen target domains. Yue et al. \cite{yue2019domain} utilize ImageNet \cite{deng2009imagenet} data to first transfer their style with a image-to-image network and then enforce consistency on feature-level with pyramid pooling between the different styles. Similarly, WildNet \cite{lee2022wildnet} transfers the style from ImageNet samples using adaptive instance normalization (AdaIN) \cite{huang2017arbitrary} and then applies contrastive learning and consistency learning using both the original and style randomized images. Peng et al. \cite{peng2021global} also apply AdaIN, but on the basis of paintings and enforce consistency between locally and globally randomized images. WEDGE \cite{kim2021wedge} proposes to use real images available in the internet to first apply a style injection and then conduct self-training with entropy-based pseudo-labels on the web-images. As the very first work for DG FSDR \cite{huang2021fsdr} operates in the frequency space and only randomizes appearance-variant frequencies while the others remain unchanged. Similar to the other works, it accesses ImageNet data for frequency style randomization. As the only work for semantic segmentation for DG AdvStyle \cite{zhong2022adversarial} does not need additional real data and therefore builds the third and smallest cluster. It searches for adversarial stylized images and afterwards augments the dataset with these newly found samples. Notably, AdvStyle conducts a small ablation study for augmentations and their adversarial stylization with Gaussian Blur and Color Jittering. But this study includes only a very small subset of the available augmentations.\\
\textbf{Augmentations}
A large variety of image augmentations can be used for the training of DNNs and one can distinguish between rule-based \cite{krizhevsky2017imagenet} and learning-based augmentations \cite{antoniou2017data}. Rule-based augmentations  include any image transformation without learnable parameters. Here we  further distinguish between geometric and color augmentations. Popular geometric augmentations are rotation, horizontal/vertical flip, scale, crops and distortions; popular color augmentations include color jitter (hue, saturation, brightness, contrast), contrast normalization, solarization, posterization. The augmentations obtained by adding different kinds of noise or blur can be summarized as a texture augmentation. More sophisticated augmentation techniques for Deep Learning were proposed in the past years like CutMix \cite{yun2019cutmix}, MixUp \cite{zhang2017mixup} and CutOut \cite{devries2017improved}. It is common practice to combine multiple different augmentations. Often, hand-crafted combinations are used but some works proposed smarter, more targeted methods to find the best augmentations. The most popular are AutoAugment \cite{cubuk2019autoaugment} and AugMix \cite{hendrycks2019augmix}. AutoAugment searches augmentation policies to improve in-domain generalization and AugMix utilizes the results of AutoAugment along with a consistency loss. However, despite these research efforts, there are no works that systematically study augmentations for out-of-domain generalization for semantic segmentation.  
\section{Method}
\subsection{General Approach}
The objective of our analysis is the systematic evaluation of augmentations for domain generalization for semantic segmentation. The number of available augmentations is large but the existing knowledge about their performance for DG is limited. For this reason, we choose a two step approach to systematically find the best performing augmentations. In the first step, we evaluated the most common but also less popular augmentations only in combination with random cropping to estimate the domain generalization capabilities for each of them separately. In the second step, we pick the five best augmentations  and combine them in a full factorial manner to analyze their interactions. In this setting, we  analyze both the in-domain and out-of-domain generalization of the augmentations. 
\subsection{Step I: Augmentation Analysis}
The choice of the augmentations to be included in the analysis is difficult. Except for the very common augmentations, not much information is available which of them supports DG. We therefore choose the augmentations in a way that balances between popular and less popular augmentations, but also between geometric and color augmentations. In table \ref{tab:augmentation_table} all augmentations included in the analysis are listed.\\
%\addtolength{\parskip}{-5mm}
\begin{table}[]
    \centering
    \begin{tabular}{c|c|c}
    \hline
         \textbf{Geometric} & \textbf{Color} & \textbf{Texture}  \\
         \hline
         Crop & Color Jitter & Gaussian Blur\\
         Hor. Flip & Brightness \& Contrast &Gaussian Noise \\
         Cutout & Grayscale & Random Snow \\
         Elastic Transform & CLAHE & Random Rain \\
         Shift-Scale-Rotate &  & Random Fog \\
          &  & Random Sun Flare \\
    \end{tabular}
    \caption{Image Augmentations included in the Step I analysis.}
    \label{tab:augmentation_table}
    \vspace{-8mm}
\end{table}
For most of the augmentations, the intensity can be modified via one or more hyperparameters. E.g.\ for color jitter, it can be defined how strong hue, saturation, brightness and contrast shall be augmented. Since it is computationally expensive to evaluate different strengths of single augmentations, we run all augmentations  in their corresponding default settings. It is possible that stronger augmentations lead to even better generalization capabilities but this is left for future work.\\
\textbf{AutoAugment \cite{cubuk2019autoaugment}} As mentioned before, AutoAugment is a popular augmentation policy search algorithm used to find the best combination of augmentation. We include AutoAugment into the Step I-Analysis because it is highly correlated to our objective. AutoAugment was developed for in-domain generalization for classification and therefore is  not directly transferable to our setting. For this reason, we choose the augmentation policy from AutoAugment for ImageNet. This policy has 25 sub-policies each of which contains two augmentations that are applied with a certain probability and a fixed magnitude. The augmentations in the subpolicies are: posterize, solarize, AutoContrast, equalize, invert, color jitter and shear. The ImageNet policy found has a clear focus on color-based augmentations which is one reason why most of these subpolicies are not included in our set defined in \ref{tab:augmentation_table}.\\
\textbf{Canny Edge Extraction}
It is shown by works like \cite{geirhos2018imagenet} that DNNs are biased towards texture. This poses a challenge for domain generalization, because the texture can strongly differ between source and unknown target domains. For this reason we propose to utilize the canny edge detection algorithm \cite{rong2014improved} as an augmentation. Based on the gradient of the image, the canny edge algorithm returns a binary image with only edge or non-edge pixels. By randomly converting a percentage $p$ of the input images into canny edge images we completely remove texture and force the network to train on the shapes of the objects, represented by the edges, only. For semantic segmentation this can be useful, because the extracted canny edges often correspond to the semantic class boundaries.  
\subsection{Step II: Full Factorial Statistical Analysis}
This study does not only aim to find a set of augmentations with good out-of-domain generalization capabilities, but also to gain more insight into the interaction between different augmentations. For this reason, we set up a full factorial design (FFD) \cite{antony2014design} for the second step. In contrast to the fractional factorial design, we have to run all possible combinations. But since there is very limited prior knowledge about the interactions of augmentations, we followed the FFD where $2^i$ experiments are necessary with $i$ being the number of impact factors; in our case the augmentations.\\
Naturally, we want to include as many augmentations as possible. However, due to computational limitations we had to restrict the number of augmentations for the full factorial analysis to $i=5$ so that we have to run $2^5=32$ experiments. \\
We use the data from the FFD experiments to fit both a linear regression model that only accounts for main effects and a quadratic regression model with first order interactions. The fitted models allow us to analyze the impact and interactions of the augmentations in a statistically valid way. %Also, as an extension using a fractional factorial design with more augmentations fitted models can be used to predict the DG capabilities of never tested sets of augmentations offering an alternative to compute intensive search algorithms like AutoAugment. %HG: Würde ich weglassen 
\subsection{Datasets}
We follow the common evaluation scheme for domain generalization for semantic segmentation known from previous works \cite{huang2021fsdr, zhong2022adversarial, peng2022semantic, choi2021robustnet}. Synthia \cite{ros2016synthia} serves as the synthetic source dataset with 9,400 images. Since no split is given from the authors, we split the dataset based on its scenes (each having 700 images) and use 8000 images for training, 700 for validation and 700 for testing. To evaluate the generalization capabilities the real world dataset Cityscapes (CS) \cite{cordts2016cityscapes} and BDD \cite{yu2020bdd100k} are used. Their training sets of 2975 and 7000 images, respectively, remain completely unused and only the validation datasets with 500 and 1000 images, respectively, are used for evaluation. The $mIoU$ is evaluated following the common practice on the 16 classes that are shared with Synthia.
\subsection{Architecture}
For all of the experiments including the FFD experiments, we employ the DeeplabV2 with a ResNet-101 backbone \cite{chen2017deeplab} as the segmentation architecture. DeeplabV2 is the de-facto standard in UDA \cite{xie2022sepico, zhang2021prototypical, tsai2018learning, Chen2019crdoco, gao2021dsp} and utilized by previous DG approaches \cite{kim2021wedge, pan2019switchable}; some of which employ the similar DeeplabV3+ network \cite{choi2021robustnet, lee2022wildnet}.\\
Only for the state-of-the-art comparison, the recent vision transformer architecture DAFormer \cite{hoyer2022daformer}, which is based on SegFormer \cite{xie2021segformer}, is used to evaluate the DG capabilities of transformer networks and show the cross-architecture applicability of augmentations.  
\subsection{Implementation Details}
As common in UDA and DG, both backbones are initialized with ImageNet \cite{deng2009imagenet} pre-trained weights. For the DeeplabV2 the SGD optimizer with a unified learning rate of $1e-3$ across all experiments and a batchsize of 20 is used. No learning rate schedule was used. The learning rate was tuned using the basic random crop augmentation. All trainings ran until convergence for 70 epochs. For the DAFormer training, the AdamW optimizer with a learning rate of $2e-4$ and a batchsize of 52 is chosen. Only 30 epochs of training were necessary. For the implementation of augmentations, the Albumentations library \cite{albumentations} and the corresponding default settings are used.
In each of the epochs the original size of the Synthia dataset was kept and the images randomly on-the-fly with a probability of $p=0.5$ augmented.\\
An important methodical question in UDA and DG is the selection of the model checkpoints for evaluation. The scientifically clean way is the selection only based on the validation performance on the source domain; in our case Synthia. However, usually better values are obtained when the best performance is directly chosen based on the target validation performance, although this method clearly contradicts the idea of domain generalization because we want to generalize to unknown target domains. Therefore, we cannot assume that we have target samples for checkpoint selection available in practice. Unfortunately, none of the existing works clearly describes how the checkpoint selection is done. For this reason and to ensure comparability, we report both values. Therefore we introduce names for two values per dataset; Cityscapes/Synthia mIoU I refers to the checkpoints where Cityscapes shows the highest performance and Cityscapes/Synthia mIoU II refers to the clean selection based on the highest performance on the source validation dataset. The Synthia mIoU I is of lower interest and therefore not reported. In the state-of-the-art comparison the clean value will be reported as the main performance value and the $mIoU$ obtained by checkpointing on the target domain is given in brackets.    
\section{Results}
\subsection{Analysis of Augmentations}
The results of the application of augmentations are shown in table \ref{tab:dg_standalone}. First, we observe that random cropping performs significantly better than resizing the input images for both in- and out-of-domain generalization. We consider resizing as the baseline where no augmentation are applied. Therefore, we select random crops as the basis augmentation which will be applied in all further experiments, since either resizing or cropping has to be applied during training due to GPU limitations. Random resized cropping performs slightly better but includes a resizing operations which means scaling of image objects. For this reason we chose the more simple random cropping to avoid too strong interactions with the basis augmentation. The second observation is that next to random cropping, none of the other augmentations achieves a significant improvement for both in- and out-of-domain generalization. Only the best performing augmentation Gaussian Blur provides $~2\%$ $mIoU$ improvement on Cityscapes. Some augmentations like cutout or sun flare lower the performance significantly. This can be expected, because augmentations can also harm the generalization if essential information are lost. Also both Canny Edge augmentation and AutoAugment do not show a beneficial effect for in- or out-of-domain generalization. Surprisingly, ColorJitter, despite being a very popular augmentation, did not cause an improvement. It is, however, possible that the basis augmentation random crop covers potential improvements from the other augmentations since it already provides a significant generalization improvement. The observation that single augmentations do not provide significant improvements leads to our step II-experiments where we combine multiple augmentations.\\
\setlength{\tabcolsep}{7pt}
\renewcommand{\arraystretch}{1.1}
\begin{table}[]
    \centering
    \begin{tabular}{c|c|c|c}
        \textbf{Augmentation} & \textbf{CS I} &  \textbf{CS II} & \textbf{Synthia II}  \\
        \hline
        \multicolumn{4}{c}{\textbf{Geometric}}\\
        \hline
         Resize& 29.6 & 29.3 & 45.9 \\
         \cline{2-4}
         RandomCrop & 35.7 & 35.4 & 60.3 \\
         \cline{2-4}
         RandomResizedCrop & 36.3 & 35.8 & 61.9 \\
         \cline{2-4}
         RCrop + Hor. Flip & 33.7 & 33.3 & 60.8 \\
         \cline{2-4}
         RCrop + Cutout & 34.3 & 33.6 & 60.7 \\
         \cline{2-4}
         RCrop + ElasticTransform & 37.7 & 36.4 & 61.8 \\
         \cline{2-4}
         RCrop + ShiftScaleRotate & 36.6 & 35.8 & 61.6 \\
         \hline
         \multicolumn{4}{c}{\textbf{Color}}\\
        \hline
         RCrop + Brightness/Contrast & 35.0 & 34.2 & 61.2 \\
         \cline{2-4}
         RCrop + ColorJitter & 35.0 & 34.2 & 61.2 \\
         \cline{2-4}
         RCrop + CLAHE & 37.9 & 35.8 & 60.9 \\
         \cline{2-4}
         RCrop + Grayscaling & 35.4 & 34.2 & 60.6 \\
         \cline{2-4}
         AutoAugment & 34.8 & 33.3 & 60.1 \\
         \hline
         \multicolumn{4}{c}{\textbf{Texture}}\\
         \hline
         RCrop + GaussNoise & 33.9 & 32.5 & 60.8 \\
         \cline{2-4}
         RCrop + GaussBlur & 37.9 & 37.0 & 60.7 \\
         \cline{2-4}
         RCrop + RandomFog & 36.6 & 36.1 & 59.8 \\
         \cline{2-4}
         RCrop + RandomRain & 36.8 & 34.8 & 60.3 \\
         \cline{2-4}
          RCrop + RandomSnow & 35.5 & 34.5 & 60.3 \\
         \cline{2-4}
         RCrop + RandomSunFlare & 34.0 & 31.7 & 60.1 \\
         \cline{2-4}
         RCrop + CannyEdge & 34.4 & 33.7 & 60.0 \\
         \hline
    \end{tabular}
    \caption{Domain Generalization for different single augmentations (next to random cropping as basis)}
    \label{tab:dg_standalone}
    \vspace{-8mm}
\end{table}
\subsection{Full Factorial Experiments}
The results of our full factorial design of experiment are listed in table \ref{tab:ffd_full}. This data is used to fit the regression models. The five best augmentations were chosen according to their performances listed in table \ref{tab:dg_standalone} for the FFD experiments. We observe that the combination of the best performing augmentations realizes significant improvements for both in- and out-of-domain generalization. The best performing combination of Elastic Transform and random resized cropping gains nearly 4\% mIoU for in-domain generalization and 2 or 4 \% improvement for out-of-domain generalization. Gaussian Blur with random resized crops performs similar but reaches no improvement on the clean selected Cityscapes mIoU II. Certain combinations affect the out-of-domain performance negatively. In particular, using all five augmentations together shows the worst of all results. Notably, a significant increase of the in-domain performance up to 63 \% mIoU is consistently correlated to a better out-of-domain generalization on the CS mIoU I. 
\begin{table}[]
\centering
\begin{tabular}{|c|c|c|c|c|c|c|c|}
\textbf{GB} & \textbf{RRain} & \textbf{ET} & \textbf{CLA} & \textbf{RRC} &\textbf{CS I} & \textbf{CS II} & \textbf{S II} \\
\hline
\hline
 X & - & - & - & -  &37.5& 36.8 &61.6 \\ 
 \hline
X & X & - & - & - & 37.3 &35.8&59.9\\ 
\hline
 X & - & X & - & - & 37.5&	36.8&62.0\\   
 \hline 
 X & - & - & X & - &38.0&	35.7&60.6\\  
 \hline 
 X & - & - & - & X &39.0&	35.5& 63.6\\
 \hline 
 X & X & X & - & - & 35.8&	34.6&60.3\\
 \hline 
 X & X & - & X & - &35.5&	33.8&59.5\\ 
 \hline 
 X & X & - & - & X & 36.3&	34.1&61.5 \\ 
 \hline
 X & - & X & X & - &37.4&	36.9&60.9\\
\hline
 X & - & X & - & X &37.6&	36.1&64.1\\
 \hline 
 X & - & - & X & X &38.2&	34.8&62.9 \\
 \hline 
 X & - & X & X & X & 38.1&33.4&63.1\\
 \hline 
 X & X & - & X & X &34.9& 33.1&61.7\\
 \hline 
 X & X & X & - & X &36.3&33.2&62.1\\
 \hline 
 X & X & X & X & - & 35.3&33.4&60.3\\
 \hline 
 X & X & X & X & X & 34.9 & 31.9&	61.6 \\
 \hline 
 - & X & - & - & - &36.4 & 35.1 & 60.6\\
 \hline 
 - & X & X & - & - & 37.1 & 36.4& 60.3\\
 \hline 
 - & X & - & X & - &36.6&	34.5&60.4\\
 \hline 
 - & X & - & - & X & 35.8&	33.7& 62.0\\
 \hline 
 - & X & X & X & - & 35.6&34.8&	60.2\\
 \hline 
 - & X & X & - & X &37.4&34.1&62.4\\
 \hline 
 - & X & - & X & X & 37.2&	33.5&62.2\\
 \hline 
 - & X & X & X & X & 36.3&32.6&61.3\\
 \hline 
 - & - & X & - & - &38.5&	36.7&62.2\\
 \hline 
 - & - & X & X & - &38.1&	36.0&61.3\\
 \hline 
 - & - & X & - & X & 39.5&	37.8&63.9\\
 \hline 
 - & - & X & X & X &38.3&35.4&64.2\\
 \hline 
 - & - & - & X & - &37.2&	36.5&61.0\\
 \hline 
 - & - & - & X & X & 38.1&	36.0&63.2\\
 \hline 
 - & - & - & - & X & 38.5&36.8&62.9\\
 \hline 
\end{tabular}
\caption[Tabelle]{FFD Experiments; GB $\rightarrow$ Gaussian Blur; CLA$\rightarrow$ CLAHE;  RRain $\rightarrow$ RandomRain; ET $\rightarrow$ Elastic Transform; RRC $\rightarrow$ Random Resized Crop ;Random Crop is used if RRC is not used}
\label{tab:ffd_full}
\vspace{-8mm}
\end{table}
\subsection{Statistical Comparison}
\begin{table}[ht]
\centering
\begin{tabular}{rrrrr}
\hline
\multicolumn{5}{c}{\cellcolor{gray!60}\textbf{In-Domain Generalization Synthia}}\\
\hline
 & Estimate & Std. Error & t value & Pr($>$$|$t$|$) \\ 
  \hline
  (Intercept) & 60.8456 & 0.2858 & 212.93 & 0.0000 \\ 
  GB & 0.2500 & 0.3195 & 0.78 & 0.4453 \\ 
  RRain & -0.6875 & 0.3195 & -2.15 & 0.0470 \\ 
  ET & 0.9850 & 0.3195 & 3.08 & 0.0071 \\ 
  CLA & -0.0100 & 0.3195 & -0.03 & 0.9754 \\ 
  RRC & \cellcolor{orange!30}2.2925 & 0.3195 & 7.18 & 0.0000 \\ 
  GB:RRain & -0.2975 & 0.2858 & -1.04 & 0.3133 \\ 
  GB:ET & -0.0475 & 0.2858 & -0.17 & 0.8701 \\ 
  GB:CLA & -0.4525 & 0.2858 & -1.58 & 0.1329 \\ 
  GB:RRC & -0.0525 & 0.2858 & -0.18 & 0.8565 \\ 
  RRain:ET & -0.5825 & 0.2858 & -2.04 & \cellcolor{orange!30}0.0584 \\ 
  RRain:CLA & \cellcolor{orange!30} 0.1775 & 0.2858 & 0.62 & 0.5432 \\ 
  RRain:RRC & -0.5725 & 0.2858 & -2.00 & \cellcolor{orange!30}0.0624 \\ 
  ET:CLA & -0.4575 & 0.2858 & -1.60 & 0.1289 \\ 
  ET:RRC & -0.1075 & 0.2858 & -0.38 & 0.7117 \\ 
  CLA:RRC & \cellcolor{orange!30} 0.0875 & 0.2858 & 0.31 & 0.7634 \\  
   \hline
\end{tabular}
\caption{Statistical Evaluation for In-Domain Generalization on Synthia. Best values are highlighted.}
\label{tab:synthia_ious}
\vspace{-4mm}
\end{table}
The statistical evaluation of the fitted quadratic model is shown in table \ref{tab:synthia_ious} and table \ref{tab:cs_ious} for in-domain and out-of-domain generalization respectively including the coefficients, standard error, t value and significance.\\
We observe that the improvements by the augmentations are comparable for in- and out-of-domain generalization with random resized crops having the highest impact and, except for gaussian blur and CLAHE, all observations are statistically significant for significance level $\alpha=5\%$. For the interactions between each pair, we can observe that most of the interactions receive a negative coefficient in the fitted model. This implies that the improvement by a combination of the two augmentations is smaller than the simple sum of the two single effects of the augmentation. It clearly indicates that there is a performance-harming interaction between most of the two augmentations. The observed behavior is reasonable and expected. When augmentations are combined their generalization increasing effects are likely to interfere so that the overall improvement is reduced, e.g. for random rain and Gaussian blur both utilize blurring, which in combination causes a negative coefficient for the interaction. Only two exceptions can be reported. The pairs random rain/CLAHE and CLAHE/random resized crop show a positive estimate but without statistical significance. However, it has to be mentioned that the statistical significance for most of the interactions is relatively low  which can be related to the small experiment size and number of data point used for the comparison. A larger statistical study with higher significance values is left for future work.\\       
\begin{table}[!ht]
\centering
\begin{tabular}{rrrrr}
  \hline
  \multicolumn{5}{c}{\cellcolor{gray!60}\textbf{Out-of-Domain Generalization Cityscapes}}\\
\hline
 & Estimate & Std. Error & t value & Pr($>$$|$t$|$) \\ 
  \hline
(Intercept) & 35.9869 & 0.4995 & 72.04 & 0.0000 \\ 
  GB & 0.3800 & 0.5585 & 0.68 & 0.5060 \\ 
  RRain & -1.1100 & 0.5585 & -1.99 & 0.0643 \\ 
  ET & \cellcolor{orange!30}1.2850 & 0.5585 & 2.30 & 0.0352 \\ 
  CLA & 0.0100 & 0.5585 & 0.02 & 0.9859 \\ 
  RRC & 0.3850 & 0.5585 & 0.69 & 0.5005 \\ 
  GB:RRain & \cellcolor{orange!30}-0.0150 & 0.4995 & -0.03 & 0.9764 \\ 
  GB:ET & -0.7075 & 0.4995 & -1.42 & 0.1758 \\ 
  GB:CLA & -0.4350 & 0.4995 & -0.87 & 0.3967 \\ 
  GB:RRC & -0.7700 & 0.4995 & -1.54 & 0.1427 \\ 
  RRain:ET & -0.5600 & 0.4995 & -1.12 & 0.2788 \\ 
  RRain:CLA & -0.2825 & 0.4995 & -0.57 & 0.5795 \\ 
  RRain:RRC & -0.9175 & 0.4995 & -1.84 & 0.0849 \\ 
  ET:CLA & -0.7850 & 0.4995 & -1.57 & 0.1356 \\ 
  ET:RRC & -0.6250 & 0.4995 & -1.25 & 0.2288 \\ 
  CLA:RRC & -0.6025 & 0.4995 & -1.21 & 0.2453 \\
   \hline
\end{tabular}
\caption{Statistical Evaluation for Out-of-Domain Generalization on Cityscapes. Best values are highlighted.}
\label{tab:cs_ious}
\vspace{-8mm}
\end{table}
The evaluation of out-of-domain generalization in table \ref{tab:cs_ious} leads to both, similar and fundamentally different, observations. Similar to in-domain-generalization, we observe that all interaction coefficients are negative; most likely for a similar reason. In contrast, the particular augmentations with a good performance are different. Elastic transform shows the best out-of-domain generalization capabilities. The combination of gaussian blur and random rain has the smallest performance drop; for in-domain random rain and CLAHE has the highest positive interaction coefficient. CLAHE and random resized crop show a positive interaction value on Synthia while it has one of the most negative interactions for out-of-domain generalization. Overall, it becomes clear that the augmentations work dissimilar for in- and out-of-domain generalization. The issue with the low significance values also exists here.

\subsection{Comparison with SOTA}
We followed the common evaluation practice of DG works in our experiments. Therefore a performance comparison with state-of-the-art approaches is of interest; especially because our augmentation-based approach is significantly easier than practically all other existing works.\\
From the comparison in table \ref{tab:sota_vergl} for the DG performance for Cityscapes, we observe that our purely augmentation-based generalization performs surprisingly competitive despite its simplicity. The clean Cityscapes mIoU II of our best performing augmentation setting already performs similar or better than DRPC, RobustNet, SW, AdvStyle and IBN. Compared with the state-of-the-art methods WEDGE, SAN+SAW and FSDR the mIoU II in brackets is in a competitive range. This is remarkable, because WEDGE and FSDR utilize real data from ImageNet which is not necessary for our approach.\\
We also evaluate the augmentations with the DAFormer \cite{hoyer2022daformer} architecture, which was not utilized for DG so far. It outperforms all existing DG approaches with only random cropping as an augmentation by a significant margin. With the best augmentations found for Cityscapes no significant further gain is obtained. However, a peak performance of 45.7\% mIoU for the Cityscapes mIoU I is obtained when color jitter is applied. That indicates that the performance of augmentations might be architecture dependent but more research is required to answer this question. The significant increase in performance is strongly correlated to the better generalization capabilities of vision transformers.

\begin{table}[ht]
\centering
    \begin{tabular}{c|c|c|c|c|}
\textbf{Method} & \textbf{} & \textbf{Synthia}\\
\hline
     && \cellcolor{blue!30}to Cityscapes\\
     \hline
     Baseline (Ours) & \multirow{13}{*}{\textbf{{ResNet-101}}} & 29.3 (29.6)\\
     
     RandomCrop (Ours) &  & 34.7 (35.6)\\
     
     IBN \cite{pan2018two} &&34.2\\ 
   
      SW \cite{pan2019switchable} && 31.6\\
      
      DRPC \cite{yue2019domain} && 37.6 \\
      
      GTR \cite{peng2021global} && 39.7\\
      
       RobustNet \cite{choi2021robustnet} & &37.2 \\
     
       FSDR \cite{huang2021fsdr} &&40.8 \\
      
      AdvStyle \cite{zhong2022adversarial} & & 37.6 \\
      
      WEDGE \cite{kim2021wedge} & & \textbf{40.9}\\
     
      SAN\&SAW \cite{peng2022semantic} & &\textbf{40.9} \\
      
      RRCrop + ET (Ours) & & 37.8 (39.5)\\
      \hline
      Baseline & \multirow{5}{*}{{\textbf{DAFormer}}}&39.6 (40.3)\\
      
      RandomCrop & &42.6\\
      RRC, ET &&42.0 (42.9)\\
      RRC, Jitter &&42.1 (45.7)\\
      RRC,Jitter,GB  &&\textbf{44.2}\\
      
      \hline
\end{tabular}
    \caption{Comparison with state-of-the-art. mIoU II in brackets.}
    \label{tab:sota_vergl}
    \vspace{-8mm}
\end{table}

\subsection{Learning Rate Dependency}
\begin{table}[]
\centering
    \begin{tabular}{c|c|c|c|c|}
     \textbf{Backbone}&$lr$&\multicolumn{3}{c|}{\textbf{Synthia}}  \\
     \hline
     \multirow{5}{*}{\textbf{DAFormer}}&&\cellcolor{blue!30} to CS&\cellcolor{yellow!30} to BDD&to Synthia\\
     
     \cline{3-5}
     &0.0007&34.4&22.5&68.8\\
     \cline{3-5}
     &0.0005&38.8&24.2&\textbf{70.6}\\
     \cline{3-5}
     &0.0001&\textbf{42.5}&\textbf{31.1}&70.1\\
     \cline{3-5}
     &0.00008&40.7&29.3&69.6\\
     \hline
    \end{tabular}
    \caption{Source and Target performances in $mIoU$ w.r.t the learning rate}
    \label{tab:learning_rate}
    \vspace{-8mm}
\end{table}

We observe that the performance on the unseen target domains shows a strong dependency on the chosen learning rate and conducted a small parameter study. The results are shown in table \ref{tab:learning_rate}. While the different learning rates have no significant impact on the performance in the source domain, the target domain performance differs substantially. For Cityscapes, the performance differs by 8.1\% mIoU and similarly for BDD by 8.6\% mIoU dependent on the $lr$. Higher learning rates clearly have a negative impact on the target domain performance.\\

\section{Discussion}
First, prior to this work, there was a lack of systematic studies about the impact and the interactions of augmentations on the training of DNNs with regard to domain generalization. This study is a first step, but more work is necessary to obtain further insight. The statistical significance has to be improved to draw more conclusions.\\
The second discussion point is the transferability of augmentations in practice. This is a crucial question, since one downside of our experiments is the high time- and computational resource consumption, which is not feasible for each new architecture or dataset. Our results indicate that different augmentations work differently well for different architectures impeding their transferability. Also the applicability across different datasets can be complicated as our in- and out-domain performance analysis has shown.\\
We reason that the ImageNet pre-trained weights of the network backbone are the explanation for the strong learning rate dependence of the target performance. Higher learning rates can lead to fast change of the initial ImageNet features causing a forgetting of the real data bias that helps improving the generalization on the target domain. This observation highlights the importance of careful hyperparameter selection and the reliance of domain generalization on the pre-trained ImageNet knowledge. Countermeasures like a learning rate schedule are left for future work.\\ 
Lastly, the topic of checkpoint selection  is crucial for domain generalization since dependent on the method applied, the performance can differ by several percent. We hope that upcoming research works follow our method of the source validation performance based selection and ensure a clean scientific standard for domain generalization benchmarking.

\section{Conclusions}
We systematically studied the effect of augmentations for in- and out-of-domain generalization for the challenging benchmark of synthetic-to-real generalization. In particular random cropping improves both in- and out-of-domain generalization and other augmentations like Gaussian blur, contrast normalization or elastic transform perform the best when being combined. In our statistical analysis, we observed that the interactions between augmentations are mostly negative for the performance and that in- and out-of-domain generalization capabilities of specific augmentations are largely dissimilar. The strong learning rate dependency of the target domain performance is highly relevant for the hyperparameter selection for UDA and DG approaches. Unexpectedly, with our best augmentation setting we obtained a performance close to the current SOTA performance showcasing that simple augmentations were underestimated for domain generalization and demanding for more research in this area.    
%\addtolength{\textheight}{-12cm}   % This command serves to balance the column lengths
                                  % on the last page of the document manually. It shortens
                                  % the textheight of the last page by a suitable amount.
                                  % This command does not take effect until the next page
                                  % so it should come on the page before the last. Make
                                  % sure that you do not shorten the textheight too much.

%%%%%%%%%%%%%%%%%%%%%%%%%%%%%%%%%%%%%%%%%%%%%%%%%%%%%%%%%%%%%%%%%%%%%%%%%%%%%%%%

%%%%%%%%%%%%%%%%%%%%%%%%%%%%%%%%%%%%%%%%%%%%%%%%%%%%%%%%%%%%%%%%%%%%%%%%%%%%%%%%

%%%%%%%%%%%%%%%%%%%%%%%%%%%%%%%%%%%%%%%%%%%%%%%%%%%%%%%%%%%%%%%%%%%%%%%%%%%%%%%%

\section*{ACKNOWLEDGMENT}

The research leading to these results is funded by the German Federal Ministry for Economic Affairs and Energy within the project “KI Delta Learning – Development of methods and tools for the efficient expansion and transformation of existing AI modules of autonomous vehicles to new domains. The authors would like to thank the consortium for the successful cooperation.”

\bibliographystyle{IEEEtran}
\bibliography{IEEEabrv,bibliography}

\end{document}